\RequirePackage[svgnames,table]{xcolor}

\documentclass[11pt,letterpaper,logo]{yalearxiv}

\pdfmapfile{+cm.map}
\pdfmapfile{+ps2pk35.map}
\pdfmapfile{+charter.map}
\pdfmapfile{+pazo.map}

\usepackage{float}
\usepackage{courier}
\usepackage{graphicx}
\usepackage{booktabs}
\usepackage{array}
\usepackage{tabularx}
\usepackage{amsmath}
\usepackage{amssymb}
\usepackage{natbib}
\setcitestyle{numbers,square,sort&compress}

\newcommand{\method}{Fisheye-VLA}

\newcommand{\setuphead}[1]{\par\medskip\noindent\textbf{#1}\enspace}

\title{Fisheye-VLA: Decoupling Coverage and Acuity\\
for Manipulation with a Single Fisheye Camera}
\runningtitle{Fisheye-VLA}

\author{%
{\normalfont\ttfamily\mdseries\fontsize{9}{11}\selectfont
\href{https://fisheyevla.github.io}{\textcolor{PaperBlue}{fisheyevla.github.io}}}\\
Ziang Ren\textsuperscript{1,2},
Zike Yan\textsuperscript{1,2},
Raymond Zhang\textsuperscript{3,4},
Xuguo He\textsuperscript{3}, and
Zhongyu Li\textsuperscript{1,2,*}\\
\textsuperscript{1}Hong Kong Embodied AI Lab\\
\textsuperscript{2}The Chinese University of Hong Kong, Hong Kong SAR, China\\
\textsuperscript{3}DeepCybo \quad
\textsuperscript{4}University of Washington, Seattle, WA, USA\\
\textsuperscript{*}Corresponding author
}

\hypersetup{
  colorlinks=true,
  linkcolor=PaperBlue,
  citecolor=PaperBlue,
  urlcolor=PaperBlue,
  pdftitle={Fisheye-VLA: Decoupling Coverage and Acuity for Manipulation with a Single Fisheye Camera},
  pdfauthor={Ziang Ren, Zike Yan, Raymond Zhang, Xuguo He, and Zhongyu Li}
}
\begin{document}

\begin{abstract}
Manipulation requires both broad scene awareness and detailed local feedback, yet conventional camera rigs provide them through separate front and wrist cameras. We present Fisheye-VLA, a visual interface that brings these capabilities together using a single passive fisheye. A global view preserves the workspace, while local perspective crops direct detail toward the interaction. The key design question is where this local visual budget should go. We answer it through a controlled re-rendering study, comparing alternative crop directions on the same recorded observations. The study finds that end-effector-centered views capture most of the estimated benefit of a much larger candidate pool, motivating a compact allocation around both hands. Our interface uses calibrated end-effector projection and motion lead to track the crops, while a shared ray encoding preserves their spatial meaning as they move. Integrated with a pretrained VLA, it achieves 84\% and 82\% success in the two expanded tabletop regions, where some target placements extend beyond the front-camera coverage, and supports shelf and conveyor manipulation. Ablations show that local crops and their viewing directions become more important in the larger workspace regions. The results demonstrate that a single fisheye can support these manipulation tasks without physical wrist cameras.

\end{abstract}

\maketitle

\begin{figure}[H]
  \centering
  \includegraphics[width=\linewidth]{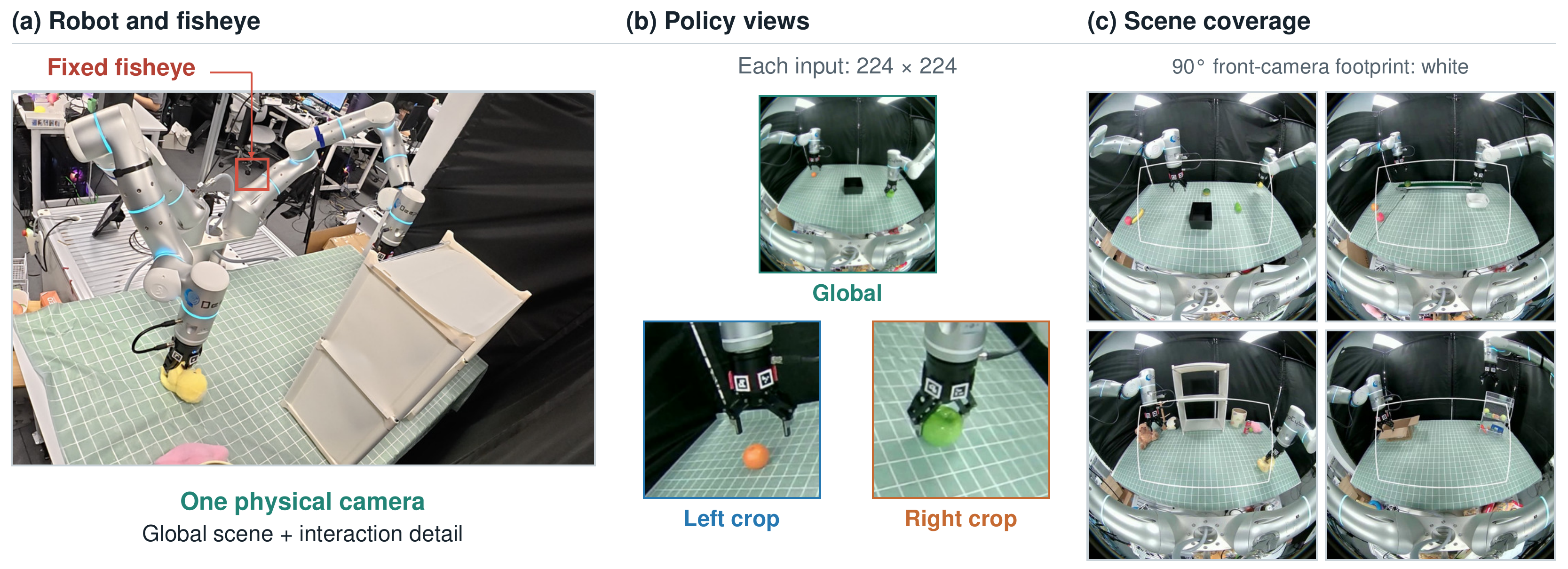}
  \caption{Fisheye-VLA uses one fixed camera (boxed) to supply a global view and two interaction-centered crops, each $224\times224$. White outlines mark the $90^\circ$ front-camera field of view. Rendering local detail from the wider fisheye image supports manipulation without wrist cameras.}
  \label{fig:teaser}\label{fig:examples}
\end{figure}

\section{Introduction}
\label{sec:intro}
A front-facing camera sees the workspace; a wrist camera sees the interaction up close. This division is a familiar and effective basis for learned manipulation~\cite{jangir2022look}. It also makes a consequential design choice: coverage and acuity are tied to separate pieces of hardware. Once the rig is built, where a policy can see and where it can see in detail are largely fixed by camera placement. We argue that this coupling is an artifact of the rig, rather than a requirement of manipulation.

The distinction matters as vision-language-action (VLA) policies take on more varied tasks~\cite{pi05}. An object at the table's edge, a mug on a shelf, and a moving object on a conveyor demand different combinations of scene context and local detail. A robot should be able to retain the whole workspace while spending detail where the current action needs it.

With a single passive fisheye, the robot records a wide field in one exposure and renders local perspective views from that image. The full scene remains available when a local view moves. We call this interface \method\ (Fig.~\ref{fig:teaser}). Its local views concentrate the policy's visual capacity around interaction using the same captured pixels. Figure~\ref{fig:teaser}c shows the coverage difference directly: the $90^\circ$ front-camera footprint omits peripheral tabletop and elevated shelf areas that remain visible in the fisheye image.

The projection explains why a fisheye is particularly suited to this role. In the ideal equidistant model, image radius is $\rho=f\theta$, so radial angular sampling density $d\rho/d\theta=f$ is constant. Equal radial angular intervals therefore occupy equal source-pixel spans near the center and periphery. A rectilinear camera instead follows $\rho=f\tan\theta$, which diverges as $\theta$ approaches $90^\circ$; a finite perspective image cannot cover a full hemisphere~\cite{kannala2006}. Local crops can therefore move over a wide angular range with comparable radial sampling. Our renderer uses the calibrated lens model to account for departures from ideal equidistance.

The sensor alone does not determine the interface. Resizing the whole fisheye image to a fixed visual budget leaves an interaction region with few pixels. Rendering every possible close-up spends tokens on redundant observations. The central question is therefore \emph{where the limited high-resolution budget should go}. Existing active-vision systems address this question through additional hardware, gaze supervision, or learned camera control~\cite{eyerobot,giava,sapave}. A calibrated fisheye gives us another option: measure the value of an allocation before introducing a controller to choose it.

Re-rendering lets us compare local views while holding the scene, robot state, and crop count fixed. The study finds that end-effector placement improves on random allocation, with a smaller additional benefit from observing the second hand. This motivates the per-hand streams developed in Sec.~\ref{sec:method} and evaluated in Sec.~\ref{sec:results}.

To implement this allocation, we need to place the crops and describe their changing viewing directions to the policy. The panorama remains available for target and scene context as the local streams follow the end-effectors. Calibrated end-effector projection places the crops, and causal motion lead keeps them ahead of a moving hand. A common spherical ray encoding tells the policy where each visual token originated as the crops move. The same rendering rule can also be driven by tracked handheld-gripper poses, making UMI-style collection a natural extension of the interface.

Our contributions are threefold. 
\textbf{Single-camera VLA manipulation over an extended workspace.} We develop a fisheye-based bimanual system and multi-task demonstration set supporting tabletop, shelf, and conveyor manipulation through one fixed sensor (Tables~\ref{tab:workspace} and~\ref{tab:tasks}).

\textbf{A shared geometric interface for global and local views.} A persistent global stream and action-centered local streams share a geometric reference, allowing coverage and local acuity to be allocated independently at the policy input (Table~\ref{tab:ablation}).

\textbf{Controlled evaluation of visual allocation.} Controlled re-rendering reveals where the extra visual budget is useful and motivates a compact end-effector prior, connecting the allocation study to a working closed-loop system (Tables~\ref{tab:placement} and~\ref{tab:closed_allocation}).



\section{Related Work}
\label{sec:related}
Camera placement determines which observations are available to a manipulation policy. We review how prior work obtains local detail and extends coverage, with the main interface differences summarized in Table~\ref{tab:interfaces}.

\begin{table}[t]
\centering
\caption{Observation interfaces and view-construction requirements.}
\label{tab:interfaces}\small
\setlength{\tabcolsep}{4pt}\renewcommand{\arraystretch}{1.12}
\begin{tabularx}{\textwidth}{@{}l*{4}{>{\raggedright\arraybackslash}X}@{}}\toprule
Interface & Supply of local detail & Allocation requirement & View cost & View rule / wrist dependence\\\midrule
Wrist + front~\cite{jangir2022look} & Physical wrist images & Camera mounting & Separate capture streams & Rig-specific wrist views\\
EyeRobot~\cite{eyerobot} & Foveated mechanical eye & Learned BC--RL gaze & Camera actuation & Eye-centered observations\\
GIAVA~\cite{giava} & Foveated active-camera images & Human gaze + learned gaze & Actuation + tokenization & Gaze-conditioned image grid\\
SaPaVe~\cite{sapave} & Actively redirected camera & Semantic camera learning & Camera actuation & Camera-conditioned policy\\
WristWorld~\cite{wristworld} & Generated wrist-view video & Geometry + video model & Reconstruction + generation & Synthesized wrist viewpoints\\
EgoMimic~\cite{egomimic} & Head and robot wrist views & Human/robot data alignment & Separate capture streams & Shared head view; robot wrists\\\midrule
\textbf{\method} & \textbf{Crops from one fisheye} & \textbf{Calibrated EE projection} & \textbf{Resampling: 1.3\,ms} & \textbf{Shared renderer; no wrists}\\\bottomrule
\end{tabularx}
\end{table}

\setuphead{Observation interfaces for VLA policies.}
OpenVLA provides an adaptable pretrained policy~\cite{openvla}, while $\pi_{0.5}$ combines flow-based action generation with heterogeneous co-training for broader generalization~\cite{pi05}. Their observation interfaces still determine which information reaches control. Hand-centric observations improve learning when they preserve task observability~\cite{hsu2022hand}, and cross-view fusion adds third-person context~\cite{jangir2022look}. Wrist input raises CLASS success from 68\% to 90\% under changing global viewpoints~\cite{class2025}; wrist views also reduce action-prediction error relative to third-person views on several visuotactile manipulation tasks~\cite{visuotactile}. We retain hand-centered detail through local fisheye views.

\setuphead{Active visual allocation.}
Kim et al. combine low-resolution peripheral vision with high-resolution gaze-centered images for precise imitation learning~\cite{kim2021gaze}.
EyeRobot learns gaze and manipulation with a mechanical eye and a BC--RL loop; its EyeGym renders eye observations from panoramic demonstrations~\cite{eyerobot}. GIAVA collects human gaze and camera actions to learn foveated processing~\cite{giava}. SaPaVe learns semantic camera control and geometry-aware execution~\cite{sapave}. AV-ALOHA adds a dedicated arm to move a stereo camera~\cite{activevision}. These systems require additional gaze supervision, control learning, or camera actuation. Human gaze offers a broader motivation: visual sampling follows the information required by the task~\cite{hayhoe2005}. We use a fixed fisheye to compare crop directions after recording, then evaluate whether end-effector positions provide a sufficient allocation rule for the tested tasks.

BFA++ learns inter-view and intra-view importance to prune supplied visual tokens~\cite{bfa}. We instead select which source pixels become local observations; view construction and token pruning are complementary.

\setuphead{Coverage beyond the current field of view.}
OmniDP uses panoramic LiDAR observations for large-workspace manipulation~\cite{omnidp}. SOMA builds persistent spatial memory using a moving head camera~\cite{soma}. We study the complementary case of persistent wide-field RGB coverage and the allocation of fine interaction detail within it.


\setuphead{Supplying local views without wrist cameras.}
WristWorld combines reconstruction and video generation to synthesize wrist views~\cite{wristworld}. Imagination at Inference synthesizes in-hand views online with pose-conditioned diffusion~\cite{imagination}. Both seek the appearance of a displaced camera, including new parallax. Our renderer inexpensively resamples observed pixels from the existing optical center. Concurrent work outlines FK-projected hand ROIs cropped from a single perspective camera~\cite{roi2026}. Such crops inherit the camera's limited field of view and offer no explicit viewing-direction cue as they move. Fisheye-VLA differs in three respects: a $220^\circ$ fisheye provides coverage beyond a rectilinear footprint, each local view is re-rendered as a calibrated perspective image rather than a pixel crop, and a shared ray encoding gives moving crops and the global view one geometric reference. We further ask \emph{where} the local budget should go through a controlled allocation study, and validate the resulting interface in closed loop over expanding workspace extents.

\setuphead{Fisheye sensing and geometric conditioning.}
Rethinking Camera Choice studies wrist-mounted fisheye sensing through spatial localization, scene generalization, and camera transfer~\cite{fisheyestudy}. VISTA considers fisheye visual grounding together with the physical feasibility of handheld trajectories~\cite{vista}. Fisheye3R identifies a geometric domain gap when perspective-trained reconstruction models encounter fisheye imagery~\cite{fisheye3r}. Our fixed-camera interface addresses local appearance through perspective rendering while retaining the full fisheye for context. For camera geometry, PRoPE uses relative positional information~\cite{prope}, and camera-conditioned imitation learning uses ray embeddings for viewpoint robustness~\cite{cameracond}. RayRoPE further conditions multi-view attention on projective ray geometry~\cite{rayrope}. For our shared-center interface, ray features give dynamic local crops and the global image a common directional reference.

\setuphead{Human demonstration interfaces.}
UMI and FastUMI collect demonstrations with handheld devices rather than robot teleoperation~\cite{umi,fastumi}. EgoMimic aligns human and robot observations for joint learning~\cite{egomimic}. Our handheld extension reuses the fisheye view-construction rule with tracked gripper poses. This preserves the global/local observation structure across collection and deployment, while robot fine-tuning supplies embodiment-specific action supervision.

\section{Method}
\label{sec:method}
\begin{figure}[!htbp]
\centering\includegraphics[width=\textwidth]{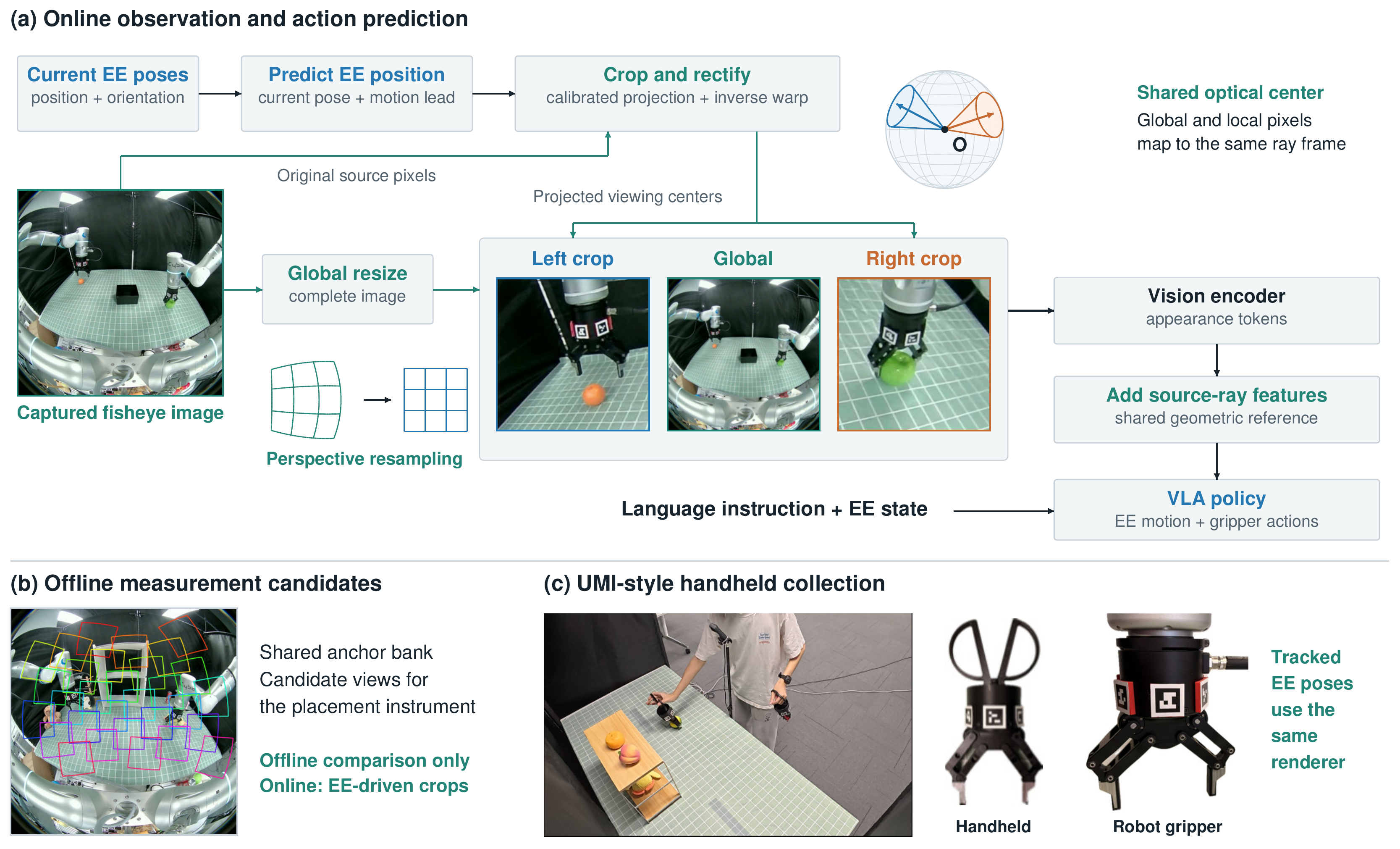}
\caption{Observation construction and policy integration. (a) Current end-effector (EE) poses and a causal velocity estimate predict the viewing centers. Calibrated projection locates these centers in the captured fisheye image; inverse perspective sampling extracts and rectifies each local view in one step. A resized global view retains the scene. The three views enter the vision encoder, and their source rays supply a common geometric reference before action prediction. The standalone sphere illustrates the shared optical center; the grid inset uses an equidistant fisheye projection. (b) Candidate-view footprints for the offline placement instrument, selected from a shared anchor bank. These are measurement candidates, not the deployed policy inputs; online crops follow the EE poses in (a). (c) Larger views of handheld collection and the matched grippers show how tracked human poses supply the same rendering input. This interface uses pose-dependent local views while preserving global coverage.}
\label{fig:pipeline}\label{fig:geometry}\label{fig:human}
\end{figure}

In this section, we construct a global/local observation interface from the crop-placement study. The policy receives a source image $I_t$, an end-effector state $s_t$ containing the current poses, and a language instruction $\ell$, and predicts an action chunk $a_{t:t+H-1}$. A global view $G_t$ preserves the complete scene, while $n$ local views $\{C_{tj}\}_{j=1}^{n}$ retain interaction detail within a fixed visual budget. Figure~\ref{fig:pipeline} connects these observations to the policy and shows how tracked handheld poses reuse the same renderer.

\subsection{Where to Place the Crops}
\label{sec:allocation_study}
Re-rendering a recorded image changes the local observation while holding the scene fixed. To compare possible allocations, we form a candidate set $\mathcal C_t$ with $K_t\leq K$ directions. Past, current, and future end-effector and instruction-referenced object positions are mapped to a shared grid of $M$ anchor directions. Future positions broaden this retrospective diagnostic search; online tracking uses only states available at the current time.

A diagnostic policy is trained with independent Bernoulli dropout over candidate views, then frozen for evaluation. For a subset $S\subseteq\mathcal C_t$, let $E(S)$ denote action-chunk endpoint error and $g(S)=E(\mathrm{global})-E(S)$ its reduction from the global-only input. Comparing EE and random placements at equal crop counts separates the value of direction from the value of adding an image. A separate object-centered condition tests whether the instruction-referenced object is a better allocation cue.

The study motivates one local stream per end-effector: useful directions concentrate near interaction, and retaining both hands avoids a separate hand-selection rule. The placement and crop-count comparisons supporting this design are reported in Sec.~\ref{sec:placement}.

\subsection{Sensor Model and Focal Rendering}
\label{sec:render}
A calibrated camera projection $\mathcal P_\kappa$ maps physical rays to source pixels, so the renderer applies independently of image resolution and lens parameterization. The captured field of view sets the available directions. We resize $I_t$ to obtain $G_t$ and render local perspective views to retain more of the captured interaction detail. Let $\eta_{tj}$ contain a virtual camera's orientation and focal parameters. For virtual intrinsics $A_\eta$ and orientation $U_\eta$, an output pixel with homogeneous coordinate $\bar u$ defines $r_\eta(u)=\mathrm{normalize}(U_\eta A_\eta^{-1}\bar u)$ in the physical camera frame. Its corresponding source pixel and rendered color are
\begin{equation}
 m_{tj}(u)=\mathcal P_\kappa\!\left(r_{\eta_{tj}}(u)\right),\qquad
 C_{tj}(u)=I_t\!\left(m_{tj}(u)\right).
\label{eq:render}
\end{equation}
Interpolation samples the source image at $m_{tj}(u)$. This inverse-warp construction performs cropping and distortion correction together, preserving a perspective image for the pretrained encoder. It also avoids resampling a globally rectified panorama before extracting the local view. Global-image rays are recovered from their source coordinates through $\mathcal P_\kappa^{-1}$, so every token can be expressed in the same camera frame.

\subsection{Focal Tracking and Adaptive Extent}
\label{sec:allocation}
The observation provides the current end-effector poses $(R_{tj},x_{tj})$ in the robot base frame. A fixed calibrated projection $\Pi_{\kappa,j}$ maps each pose to an image center, $p_{tj}=\Pi_{\kappa,j}(R_{tj},x_{tj})$. The mapping is fitted offline from reference image observations paired with recorded end-effector poses. The renderer and policy both use these Cartesian states directly; crop placement requires no object detector.

A crop centered on the current hand can leave little detail ahead of a reach. Anticipatory gaze during manipulation motivates looking ahead of the current pose~\cite{land1999}. We predict $\hat x_{tj}=x_{tj}+\tau v_{tj}$, where $v_{tj}$ is a causal backward-difference velocity and $\tau$ is the lead interval. Projecting $(R_{tj},\hat x_{tj})$ sets the local viewing direction while retaining the current orientation. The virtual optical axis follows the source ray recovered at that projected center through $\mathcal P_\kappa^{-1}$, connecting pose tracking to the sampling map in Eq.~\ref{eq:render}.

Faster motion makes the predicted center less precise, so the viewport expands with speed. With reference speed $v_{\mathrm{ref}}$ and expansion coefficient $\beta$, we use $\alpha_{tj}=1+\beta\,\mathrm{clip}(\|v_{tj}\|/v_{\mathrm{ref}},0,1)$ and local focal scale $f_{tj}=f_{0,tj}/\sqrt{\alpha_{tj}}$. The baseline focal scale $f_{0,tj}$ specifies the nominal extent; the motion term widens it during a reach.

\subsection{Spherical Ray Positional Encoding}
\label{sec:geometry}
The same local patch coordinate can refer to different physical directions as a crop moves. To preserve this spatial identity, we attach its source ray $r$ to the visual token $h$:
\begin{equation}
 \widetilde h=h+e(r),\qquad e(r)=W_2\,\sigma\!\left(W_1\gamma(r)\right).
\label{eq:ray}
\end{equation}
The function $\gamma$ combines the unit ray with Fourier features over $L$ frequency bands. A shared network embeds rays from the global and local views, and its output projection is initialized to zero to preserve the pretrained features at initialization. Matching directions receive matching ray features even when their image coordinates differ.

\subsection{Policy Integration and Deployment}
\label{sec:policy}
The global view and local crops enter the VLA's existing image pathways. The vision encoder produces appearance features, the ray residual supplies their viewing directions, and the policy conditions on these tokens, $s_t$, and $\ell$ to predict $a_{t:t+H-1}$. The robot executes motion and gripper commands from this chunk. Training and deployment use the same view construction, so the policy learns to interpret the moving local inputs it will receive online.



\subsection{UMI-Style Human Data Collection}
\label{sec:collection}
We extend the interface to human demonstrations using the fixed fisheye camera and a mechanically matched handheld gripper. The wide field of view accommodates large hand movements, reducing tracking loss caused by hands leaving the image. We render auxiliary perspective views from the fisheye recordings for fiducial detection and gripper-pose estimation, while image-based recognition estimates the gripper's open/closed state. The recovered poses drive the same global/local renderer used for robot observations (Fig.~\ref{fig:human}c), without requiring cameras on the handheld grippers. Keeping the table setup and camera placement consistent with robot deployment reduces the visual and spatial gap between the two data sources.

Short tracking gaps are interpolated, and trajectories are smoothed and aligned with video frames before motion targets are computed. Human pretraining is followed by robot fine-tuning, with domain-specific action normalization accommodating differences in motion statistics. Future extensions could use a tabletop calibration board to recover absolute gripper motion in a shared coordinate frame, reducing dependence on a matched collection setup.
\FloatBarrier

\section{Experimental Setup}
\label{sec:experiments}\label{sec:setup}
The evaluation tests whether the allocation rule improves manipulation, whether its gains persist as the workspace expands, and whether the renderer supports both robot and handheld demonstrations. We separate the frozen-policy diagnostic from independently trained closed-loop policies to distinguish observation effects from learned control performance.

\setuphead{Platform and training.}
The robot uses two seven-axis Flexiv Rizon 4s arms with GN01 grippers and one fixed head-mounted $220^\circ$ fisheye. The ordinary front-camera baseline has a $90^\circ$ field of view, illustrated by the white outlines in Fig.~\ref{fig:teaser}c. Fisheye source images are $1944\times1944$, and robot images and trajectories are sampled at 30\,Hz. Policies start from $\pi_{0.5}$~\cite{pi05} and predict 30-step chunks of 14-dimensional actions, with an end-effector delta and a gripper command per arm. Cartesian impedance control executes trajectories at 30\,Hz. Each sensing configuration uses the same 300 teleoperated demonstrations per task and 20,000 training steps, on one NVIDIA RTX PRO 6000 with batch size 12. Random, object-centered, and EE crop policies are trained separately using their deployment crop rules and this shared budget.

\setuphead{Rendering and token budget.}
The global image and each local view are $224\times224$, with bilinear sampling. Calibration uses the OpenCV fisheye model and fixed per-arm pose-to-image mappings. We set $\tau=0.5$\,s, $\beta=0.5$, and $v_{\mathrm{ref}}=0.204$\,m/s, the empirical high-percentile speed. The local area therefore expands by at most $1.5\times$. Ray encoding uses $L=4$ Fourier bands. The encoder's $14\times14$ patches produce 256 tokens per image: global plus two local views and the front camera plus two wrist cameras each supply 768 visual tokens. Ray features do not add tokens. Median deployment latency on an RTX 5090 is 87\,ms, including 1.3\,ms for rendering.

\setuphead{Placement diagnostic.}
We evaluate 1,500 held-out frames from 120 episodes. Candidate directions are snapped to a shared bank of $M=30$ anchors, yielding at most $K=16$ candidates per window. Figure~\ref{fig:pipeline}b shows the source-image footprints of these anchor views used for the offline diagnostic. Conditions are paired using the same source frame and inference randomness within each packed batch. We report endpoint-error reductions in millimeters with episode-clustered 95\% bootstrap confidence intervals. Round~1 provides the EE--random contrast, while Round~2 provides the crop-count and object-centered contrasts in Table~\ref{tab:placement}.

\begin{figure}[!htbp]
\centering\includegraphics[width=\textwidth]{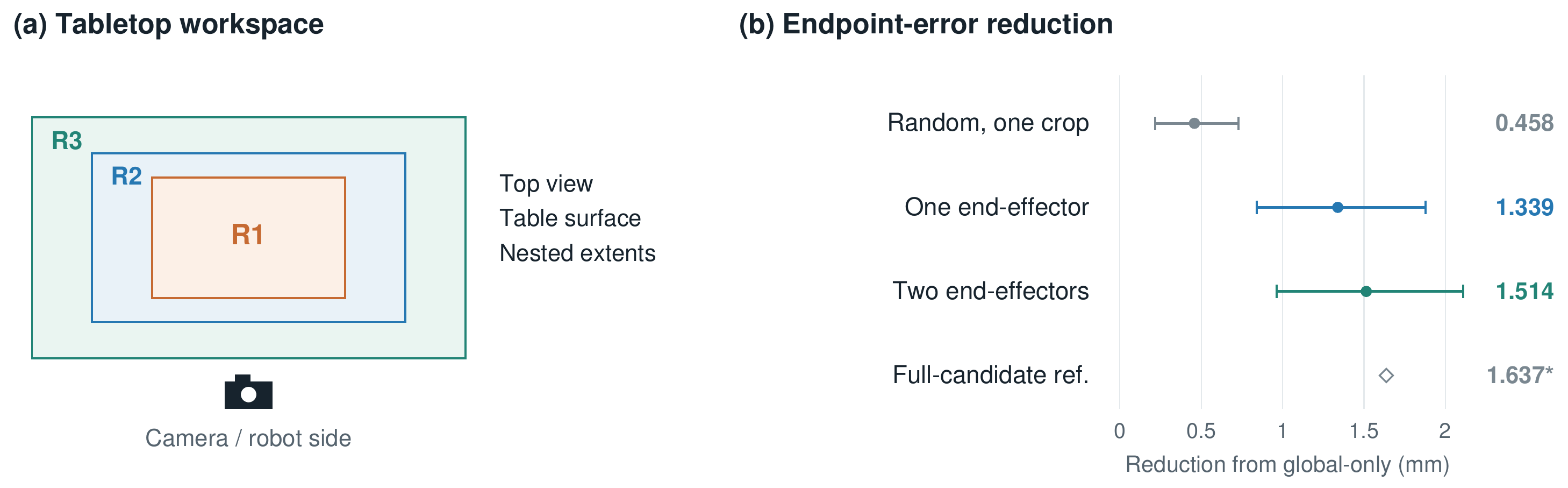}
\caption{Evaluation geometry and crop-placement gains. (a) Top view of the tabletop: R1--R3 denote increasingly large placement extents, with the camera on the robot side. (b) Round~2 endpoint-error reductions relative to the global-only input, with 95\% episode-clustered intervals; the diamond is the derived full-candidate reference. EE-centered views account for most of the estimated reduction.}
\label{fig:gains}
\end{figure}

\begin{figure}[!htbp]
\centering\includegraphics[width=\textwidth]{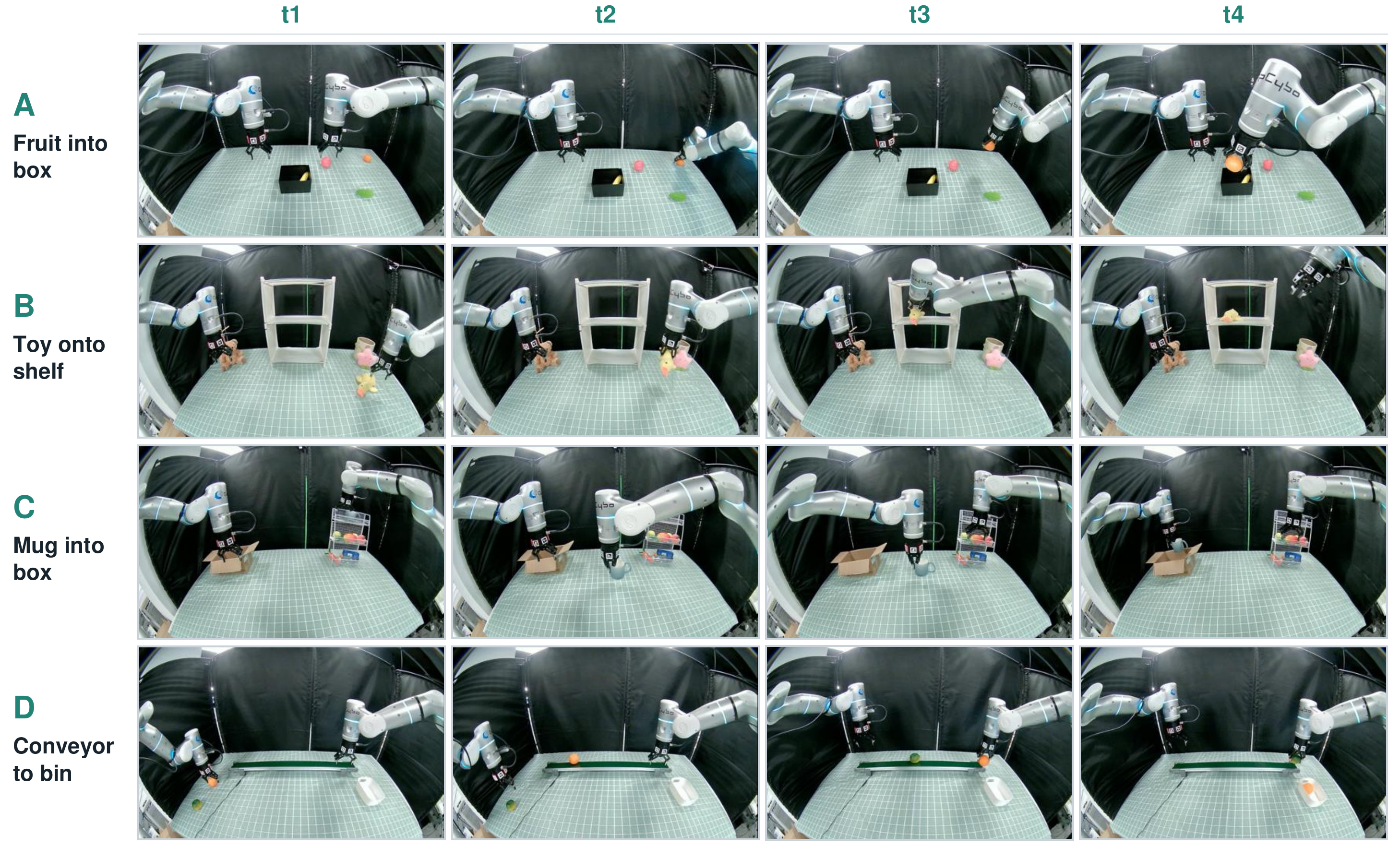}
\caption{Recorded execution sequences for the four task families. Columns show ordered stages from approach through transfer to completion; the same global/local rendering rule and multi-task policy serve every row. Shelf interaction requires keeping the elevated destination in context, and conveyor transfer requires following a moving object. The sequences illustrate reuse of the interface across different spatial and timing demands. Frame spacing indicates order, not equal elapsed time.}
\label{fig:tasks}
\end{figure}

\setuphead{Tasks and spatial protocol.}
Task A transfers an instructed fruit from ten categories into a box. Task B places a duck toy onto a shelf, and Task C transfers a mug from the shelf into a box. Task D transfers fruit between a conveyor and a bin in both directions. Success requires completing the transfer and leaving the object at its destination without dropping it. These task families test target selection, elevated placement, retrieval, and moving-object interaction (Fig.~\ref{fig:tasks}).

Task A uses predefined placement grids in nested tabletop extents: R1 is $80\times50$\,cm, R2 is $130\times70$\,cm, and R3 is $180\times100$\,cm (Fig.~\ref{fig:gains}a). Each configuration--region condition has 50 executed trials, and the table reports their observed success rates and arithmetic regional mean. Some target placements in R2 and R3 extend beyond the coverage of the front or wrist cameras; these conditions are evaluated with the same completion criterion as R1.


\setuphead{Handheld transfer.}
A separate fruit-to-shelf task tests pretraining with 300 human demonstrations before fine-tuning on 50 or 100 robot demonstrations. The collection device is a thumb-operated, GN01-shaped gripper with six AprilTags; its body can be 3D-printed at a material cost of approximately US\$1. A single fisheye image stream provides both visual observations and the inputs for gripper-pose and open/closed-state estimation, without additional cameras or motion-capture equipment. AprilTag detection in rectified views supplies poses for the shared renderer. The tripod-mounted fisheye matches the nominal robot camera placement, and each training condition is evaluated over 50 trials.
\FloatBarrier

\section{Results}
\label{sec:results}
We first assess crop placement, then compare camera configurations and interface components on the same regional scale. Task-suite and handheld-transfer results test whether the observation rule remains useful beyond the tabletop allocation experiment.

\subsection{Crop Placement}
\label{sec:placement}
At matched crop counts, an EE-centered view improves endpoint prediction over a random direction by 0.890\,mm [0.520, 1.297] in Round~1 (Table~\ref{tab:placement}). Round~2 gives reductions of 0.458\,mm for a random crop and 1.339\,mm for one EE crop, a point-estimate difference of 0.881\,mm (Fig.~\ref{fig:gains}). The placement change contributes more than simply adding a randomly directed local image.

\begin{table}[b]
\centering\caption{Offline placement: paired gains with 95\% CI.}
\label{tab:placement}\footnotesize
\setlength{\tabcolsep}{3pt}\renewcommand{\arraystretch}{1.08}
\begin{tabular}{@{}lcrl@{}}\toprule
Contrast & Round & $\Delta$ (mm) & 95\% CI\\\midrule
EE $-$ random, one each & 1 & 0.890 & [0.520, 1.297]\\
Two EE $-$ one EE & 2 & 0.175 & [0.007, 0.346]\\
All candidates $-$ one EE & 2 & 0.298 & [$-$0.034, 0.626]\\
Object $-$ random, one & 2 & $-$0.148 & [$-$0.386, 0.126]\\
Object $-$ random, three & 2 & 0.030 & [$-$0.298, 0.360]\\\bottomrule
\end{tabular}
\end{table}

A second EE crop adds $0.175\mathrm{mm}$ $[0.007,0.346]$, bringing the reduction to $1.514\mathrm{mm}$. Adding the measured all-minus-one increment to the one-EE gain yields a $1.637\mathrm{mm}$ full-candidate reference. The two-crop gain therefore corresponds to approximately $92.5\%$ of this reference, a descriptive point estimate within the evaluated candidate pool. Separately, the Round~1 single-crop retention estimate is $81.9\%$, with a $95\%$ confidence interval of $[66.3\%,102.5\%]$.

The oracle's best direction falls within $20^\circ$ of an EE in 69.9\% of windows [61.3, 78.3], against a 28.4\% chance rate computed from each window's candidate distribution. Hit/chance rates of 30.1/8.1\% at $10^\circ$ and 78.8/47.6\% at $30^\circ$ show positive paired enrichment at each scale. Thus the EE concentration extends beyond a single cone threshold.

Motion alone is insufficient to choose a useful hand when both hands are stationary. In 37 such windows from 35 episodes, an independently evaluated oracle still obtains 2.427\,mm [1.318, 3.825] of gain. Retaining one view per hand avoids this selection step. In the $0$--$30^\circ$ off-axis subset, exact-ray placement reduces endpoint error by a further 0.293\,mm [0.011, 0.622] relative to anchor-snapped placement.

The closed-loop crop block in Table~\ref{tab:workspace} evaluates independently trained policies with two local views each. EE crops achieve a regional mean of 79.3\%, compared with 33.3\% for random and 45.3\% for object-centered crops. EE placement leads in every region. Object-centered placement improves over random in closed loop, although its offline contrasts include zero: the frozen diagnostic evaluates endpoint prediction, whereas the separately trained policies must complete entire tasks.

\subsection{Workspace Coverage and Ablations}
\label{sec:workspace}\label{sec:ablation}
Table~\ref{tab:workspace} puts sensor configurations, component removals, and crop placements on the same Task A evaluation. The rig with a front camera and two wrist cameras is strongest in R1, at 92\% against our 72\%. Its success decreases to 58\% in R2, where our interface reaches 84\%. In R3, its success falls to 14\%, while Fisheye-VLA succeeds in 82\% of trials. Some placements in the expanded regions extend beyond the ordinary camera views, contributing to their lower success. The front camera alone reaches 44\%, 38\%, and 4\%, and two wrist cameras reach 76\%, 36\%, and 0\%.

\begin{table}[t]
\centering\caption{Task A: regional success and mean (\%).}
\label{tab:workspace}\label{tab:ablation}\label{tab:closed_allocation}\small
\setlength{\tabcolsep}{3pt}\renewcommand{\arraystretch}{1.07}
\begin{tabular}{@{}lrrrr@{}}\toprule
Configuration & R1 & R2 & R3 & Avg.\\\midrule
\multicolumn{5}{@{}l}{\textit{Camera baselines}}\\
Global fisheye (w/o crops) & 68 & 36 & 30 & 44.7\\
Front camera + two wrists & \textbf{92} & 58 & 14 & 54.7\\
Front camera & 44 & 38 & 4 & 28.7\\
Two wrist cameras & 76 & 36 & 0 & 37.3\\
Fisheye + two wrists & 70 & 74 & 78 & 74.0\\\midrule
\multicolumn{5}{@{}l}{\textit{Interface ablations}}\\
w/o motion lead & 68 & 76 & 68 & 70.7\\
w/o ray encoding & 62 & 64 & 40 & 55.3\\\midrule
\multicolumn{5}{@{}l}{\textit{Crop placement}}\\
Random crops & 42 & 30 & 28 & 33.3\\
Object-centered crops & 46 & 48 & 42 & 45.3\\\midrule
\textbf{Fisheye-VLA} & 72 & \textbf{84} & \textbf{82} & \textbf{79.3}\\\bottomrule
\end{tabular}
\par\vspace{2pt}\parbox{\columnwidth}{\scriptsize Every cell has 50 trials; Avg. is the mean over R1--R3. Global fisheye is also the no-focal-crop ablation.}
\end{table}

The full interface remains competitive with fisheye plus two physical wrists, whose regional rates are 70\%, 74\%, and 78\%. Rendered crops retain a common viewpoint and deterministic geometry while removing the wrist image streams. The observed differences are modest; the comparison supports a simpler sensing interface with similar task performance.

Keeping only the global image reduces R3 success from 82\% to 30\%, while the reduction in R1 is four points. The source still covers the target, but uniform resizing allocates little detail to its interaction region. Local crops both retain more local pixels and rectify the perspective view. Fisheye3R reports a mismatch when perspective-trained models receive fisheye imagery~\cite{fisheye3r}; a related appearance mismatch could contribute here. These coupled changes explain why this ablation tests the entire local-view branch.

Removing ray encoding reduces R3 success by 42 points, compared with ten points in R1. Larger workspace extents expose the policy to a broader range of crop directions, making explicit source rays more useful. Removing motion lead gives smaller decreases of four, eight, and fourteen points. The observed trend favors leaving visual room ahead of a reach, while the crop and geometry ablations account for the largest outer-region losses.

\subsection{Task Suite}
\label{sec:tasks}
The same multi-task policy completes tabletop transfer, shelf placement and retrieval, and conveyor transfer (Table~\ref{tab:tasks}). Shelf tasks require retaining a destination away from the central table while resolving the grasp or release. Conveyor transfer adds a timing requirement because the object moves during approach. Both conveyor directions have lower success than the static shelf tasks, consistent with this additional demand on tracking and control.

\begin{table}[t]
\centering\caption{Four task families; Task D is evaluated in both directions.}
\label{tab:tasks}\small
\setlength{\tabcolsep}{4pt}
\begin{tabular}{@{}llr@{}}\toprule
 & Task & Success (\%)\\\midrule
A & Instructed fruit into box & 79\\
B & Duck toy onto shelf & 88\\
C & Mug from shelf into box & 86\\
D & Conveyor $\rightarrow$ bin & 72\\
D$'$ & Bin $\rightarrow$ conveyor & 68\\\bottomrule
\end{tabular}
\par\vspace{2pt}\parbox{\columnwidth}{\scriptsize Task A is the mean over three regions; each task--region condition has 50 trials.}
\end{table}

\subsection{Human Data}
\label{sec:human}
Handheld pretraining transfers the renderer to a separate fruit-to-shelf task (Table~\ref{tab:human_counts}). At the 100-robot-demonstration budget, 300 human demonstrations improve success from 46\% to 72\%. At the 50-demonstration budget, the observed rates are 36\% and 48\%. The larger gain at the higher fine-tuning budget suggests that robot demonstrations help convert the pretrained motion representation into reliable gripper control.

\begin{table}[t]
\centering\caption{Handheld pretraining for fruit-to-shelf.}
\label{tab:human_counts}\small
\setlength{\tabcolsep}{5pt}
\begin{tabular}{@{}rrr@{}}\toprule
Human demos & Robot demos & Success (\%)\\\midrule
0 & 50 & 36\\
300 & 50 & 48\\
0 & 100 & 46\\
300 & 100 & 72\\\bottomrule
\end{tabular}
\par\vspace{2pt}\parbox{\columnwidth}{\scriptsize Each condition has 50 trials.}
\end{table}

\section{Discussion}
\label{sec:discussion}
The placement study and closed-loop experiments support an end-effector prior for the tasks considered here. Retaining one local view per hand concentrates detail around interaction and avoids choosing a hand from motion alone. The region comparisons further show why coverage alone is insufficient: a global fisheye keeps the target visible, yet performs substantially worse without the local crops.

Shared-center crops cannot remove occlusion, and pose projection depends on calibration. Human pretraining also adds both data and optimization. Broader task evaluation and calibration transfer are natural next steps for this single-fisheye interface.

\section{Conclusion}
We presented \method, a single-fisheye observation interface for bimanual VLA manipulation. A controlled re-rendering study motivated one local perspective view per end-effector, and the deployed interface tracks these views through calibrated end-effector projection. The shared ray encoding lets the policy use their changing viewing directions. Across the tested tasks, this input supports manipulation over a wider workspace than the rig with a front camera and two wrist cameras, using the same visual token budget. The handheld experiments also show that the renderer can be reused for demonstration collection. These results provide a practical basis for replacing a multi-camera rig with local views rendered from one fisheye.

\section*{Acknowledgment}
This work was in part supported by the InnoHK initiative of the Innovation and Technology Commission of the Hong Kong Special Administrative Region Government via the Hong Kong Centre for Logistics Robotics. We also thank DeepCybo for its support, and Hualong Liu, Peize Li, and Zhekai Wang for their assistance.

\begingroup
\footnotesize
\setlength{\bibsep}{2pt plus 0.4ex}
\bibliographystyle{unsrtnat}
\bibliography{references/references}
\endgroup

\end{document}